\documentclass{article}
\usepackage{amssymb,amsthm,amsmath}
\newcommand{\EE}{\mathbb{E}}

\newcommand{\PP}{\mathbb{P}}
\newcommand{\RR}{\mathbb{R}}
\newcommand{\LL}{{\cal L}}
\newcommand{\XX}{{\cal X}}
\newcommand{\YY}{{\cal Y}}
\newcommand{\indic}{{\bf 1}}

\newcommand{\als}[1]{ \begin{align*} #1  \end{align*}}
\newcommand{\eqs}[1]{ \begin{equation*} #1  \end{equation*}}

\newtheorem{corollary}{Corollary}
\newtheorem{definition}{Definition}
\newtheorem{proposition}{Proposition}

\newtheorem{example}{Example}
\begin{document}

\title{An extension of Mcdiarmid's inequality}

\author{Richard Combes \thanks{Universit\'e Paris-Saclay, CNRS, CentraleSup\'elec, Laboratoire des signaux et syst\`emes, France}}

\maketitle

\begin{abstract}
We generalize McDiarmid's inequality for functions with bounded differences on a high probability set, using an extension argument. Those functions concentrate around their conditional expectations. We illustrate the usefulness of this generalized inequality on a few examples. We further extend the results to concentration in general metric spaces.
\end{abstract}

\section{Introduction}
Consider $\XX \subset \RR^n$ a set and $X = (X_1,...,X_n) \in \XX$ a random vector with independent entries. We say that a function $f:\XX \to \RR$ has $c$-bounded differences with $c \in (\RR^{+})^n$ if its value can change by at most $c_i$ when its $i$-th entry is modified.
\begin{definition}[bounded differences]
	 Consider a vector $c \in (\RR^{+})^n$ and a function $f:\XX \to \RR$. $f$ has $c$-bounded differences if and only if $|f(x) - f(y)| \leq c_i$ for all $(x,y) \in \XX^2$ such that $x_j = y_j$ for all $j \neq i$.
\end{definition}
McDiarmid's inequality states that if $f$ has $c$-bounded differences, then $f(X)$ concentrates around its expected value. 	
\begin{proposition}[\hspace{1sp}\cite{mcdiarmid}]
If $f$ has $c$-bounded differences on $\XX$, then for all $\epsilon \geq 0$: 
$$
\PP[f(X) - \EE[f(X)] \geq \epsilon ] \leq \exp\left( - \frac{2 \epsilon^2}{\sum_{i=1}^n c_i^2} \right).
$$
\end{proposition}
In the present work we consider the case where the finite differences property only holds on a high probability subset $\YY \subset \XX$, so that $f$ has "bounded differences with high probability". The behaviour of $f$ outside of $\YY$ can be arbitrary.
\begin{definition}[bounded differences on a subset]
	 Consider a set $\YY \subset \XX$, a vector $c \in (\RR^{+})^n$ and a function $f:\XX \to \RR$. $f$ has $c$-bounded differences on subset $\YY$ if and only if $|f(x) - f(y)| \leq c_i$ for all $(x,y) \in \YY^2$ such that $x_j = y_j$ for all $j \neq i$.
\end{definition}
If $f$ has bounded differences on a high probability subset, it may not concentrate around its expected value as shown by the following example.
\begin{example}
	Consider $(X_1,...,X_n)$ i.i.d. Bernoulli($1/2$), $\YY = \{0,1\}^n \setminus \{(1,...,1) \}$ and $f(X) = 2^n  \indic\{ X \not\in \YY\}$. $f$ has bounded differences over subset $\YY$, whose probability is high $\PP[X \in \YY] = 1 - 2^{-n}$. However $f$ concentrates around $\EE[f(X) | X \in \YY] = 0$, not $\EE[f(X)]  = 1$.  
\end{example}
This example suggests that $f(X)$ should concentrate around its conditional expectation, which is correct, but not straightforward. Indeed one cannot upper bound $\PP[ f(X) - \EE[f(X)] \geq \epsilon  | X \in \YY]$ with McDiarmid's inequality since in general $(X_1,...,X_n)$ are not independent conditional to $X \in \YY$, unless $\YY$ has a specific structure. 

\section{Related work}
The strength of McDiarmid's inequality lies in its applicability (see \cite{mcdiarmid2} for an extensive survey): set $\XX$ may be completely arbitrary, and, even when $f$ is involved, it is usually easy to check that the bounded differences assumption holds. Two notable applications are combinatorics and learning theory. Two representative results are the concentration of the chromatic number of Erdos-R\'enyi graphs~\cite{bollobas}, and the fact that stable algorithms have good generalization performance~\cite{bousquet}. Namely, if the output of a learning algorithm does not change too much when a single training example is modified, then it performs well on an unseen example. An overview of how McDiarmid's inequality can be applied to several problems in information theory is found in \cite{raginsky}. For instance, \cite{liu} applies McDiarmid's inequality to solve the problem of channel resolvability. 

Motivated by the study of random graphs, \cite{kim,schudy,vu,vu2} have provided concentration inequalities for particular classes of functions $f$ (e.g. polynomials) which have bounded differences with high probability. Indeed the number of subgraphs such as triangles or cliques can be written as a random polynomial in the entries of the adjacency matrix. 

On the other hand, concentration inequalities for general functions whose differences are bounded with high probability were provided in \cite{kutin,warmke} and \cite{kontorovich} considered a more general case where differences are subgaussian and may be unbounded. In a nutshell,  \cite{kutin,warmke} assume that there exists vectors $b$ and $c$, with $b_i \geq c_i$ for all $i$ such that function $f$ has $c$-bounded differences on $\YY$ and $b$-bounded differences on $\XX$.  The provided concentration inequalities usually give a strong improvement over McDiarmid's inequality, but
are not informative if $b$ is too large. Our results shows that this is an artefact, since all the required "information" about the behaviour of $f$ outside of $\YY$ is contained in $\PP[X \not\in \YY]$. A toy example of this phenomenon is found below.
\begin{example}
Consider $(X_1,...,X_n)$ i.i.d. Bernoulli($1/2$), $\YY = \XX \setminus \{(0,...,0), (1,...,1)\}$, $B \geq 0$  and 
\eqs{f(X) = \begin{cases} +B & \text{ if  } X = (0,...,0) \\
-B & \text{ if } X = (1,...,1) \\
\frac{1}{n} \sum_{i=1}^n 2(X_i-1) & \text{ otherwise.} 
\end{cases}.}
It is noted that $p=\PP[X \not\in \YY] = 2^{1-n}$. As shown in the next section, for all $B \geq 0$, our results guarantee that 
$$\PP[ f(X) \geq \epsilon + 2^{1-n}] \leq 2^{-n} + \exp( -2 n \epsilon^2 )$$
while other inequalities become vacuous for $B$ arbitrarily large. 
\end{example}

\section{Generalized McDiarmid's Inequality}
\subsection{Statement}
Proposition~\ref{prop:main} shows that if $f$ has $c$-bounded differences on a high probability subset, then it concentrates around its conditional expectation.
\begin{proposition} \label{prop:main}
Consider $f$ with $c$-bounded differences over a subset $\YY$, then for all $\epsilon \geq 0$:
$$\PP[ f(X) - \EE[f(X)|X \in \YY] \geq \epsilon + p \bar{c}] \leq p  + \exp\left(  - {2 \epsilon^2 \over \sum_{i=1}^n c_i^2 }  \right).$$ 
where $\bar{c} = \sum_{i=1}^n c_i$ and $p=\PP[X \not\in \YY]$.
\end{proposition}
Some remarks are in order:
\begin{itemize}
 \item[(i)] In some situations, $\PP[X \not\in \YY]$ will be exponentially small, while $\bar{c}$ will be independent of $n$, so that $p = a^{-n}$ and $c_i = b/n$ for all $i$ for some $a,b > 0$. In that case one obtains the same exponent as in McDiarmid's inequality. Think for instance of the case where $f(X) = (1/n) \sum_{i=1}^n X_i$ for all $X \in \YY$. 
\item[(ii)] If $f$ has $c$-bounded differences over $\YY$, then it also does over any $\YY' \subset \YY$, at the cost of increasing $p$. If $p$ is controlled by another concentration inequality that holds for some family of sets, one can optimize the bound over $p$ to obtain a refined inequlity.
 \item[(iii)] The behavior of $f$ outside of $\YY$ may be arbitrary, and in particular $f$ may even be unbounded so that $\sup_{x \in \XX} f(x) = +\infty$. It is also noted that if $f$ is bounded, the difference between the expectation and conditional expectation can be controlled in a simple manner by:
$$
	| \EE[f(X)] - \EE[f(X)| X \in \YY] | \le  {p \over 1-p} \sup_{x \in \XX} |f(X)|
$$
\end{itemize}
The proof relies on an extension argument from~\cite{mcshane}, similar to~\cite{kirszbraun}. The idea is apply McDiarmid's inequality to $g$ which is a "smoothed" version of $f$ that has $c$-bounded differences and that is equal to $f$ over subset $\YY$.
\subsection{Proof}
To prove the result we first use the fact that $f$ has $c$-bounded differences on subset $\YY$ if and only if it is $1$-Lipschitz over $\YY$ with respect to distance $d_c$
$$
|f(x) - f(y)| \leq d_c(x,y) \quad \hfill \forall (x,y) \in \YY^2
$$
with $d_c$ the weighted Hamming distance
$$
d_c(x,y) = \sum_{i=1}^n c_i \indic(x_i \ne y_i) 
$$
We then use an extension argument from \cite{mcshane}, let
$$
g(x) = \inf_{y \in \YY} \{  f(y) + d_c(x,y) \}.
$$
Then $g$ is a Lipschitz extension of $f$: $g$ is $1$-Lipschitz over $\XX$ with respect to $d_c$, and $g(x) = f(x)$ for all $x \in \YY$. To derive the inequality decompose:
\als{
\PP[  f(X) - \EE[g(X)] \geq \epsilon] 
& \leq \PP[ f(X) - \EE[g(X)] \geq \epsilon , X \in \YY] \\ & + \PP[X \not\in \YY]
}
Since $g$ has $c$-bounded differences and equals $f$ on $\YY$:
\als{
\PP[ f(X) - \EE[g(X)] \geq \epsilon , X \in \YY] &\le \PP[g(X) - \EE[g(X)] \geq \epsilon] \\
&\le \exp\left( - \frac{2 \epsilon^2}{\sum_{i=1}^n c_i^2} \right)
}
To control the expectation of $g$ we use the fact that:
\als{
	g(X) &= \inf_{y \in \YY} \{  f(y) + d_c(X,y) \} \\ &\le \begin{cases} f(X) & \text{ if } X \in \YY \\ \EE[f(X) | X \in \YY ] + \bar{c} & \text{ if } X \not\in \YY \end{cases}
}
since 
$$\sup_{(x,y) \in \XX^2} d_c(x,y) = \bar{c}$$
and 
$$\inf_{y \in \YY} f(y) \le \EE[f(X) | X \in \YY ]$$ 
Taking expectations we get
\begin{align*}
	\EE[g(X)] &\le \EE[ f(X) \indic( X \in \YY )] \\
	&+ \EE[ (\EE[f(X) |X \in \YY] + \bar{c})\indic( X \not\in \YY )] \\
	&= \EE[ f(X) | X  \in \YY] + p \bar{c}
\end{align*}
Substituting $\EE[g(X)]$ by this upper bound yields the result.

\section{Examples}

In this section we cover a few illustrative examples. In those examples, McDiarmid's inequality can be applied but the bounds it gives are relatively loose (or vacuous) but the generalized inequality provides a more accurate control. Also, in all those examples, applying the generalized version of McDiarmid's inequality is not significantly more complex than its basic version. 

\subsection{Triangles in a Random Graph}

Our first illustrative example is the concentration of the number of triangles in random graphs. Consider a random graph over $m$ nodes, represented by $X = (X_{(i,j)})_{1 \le i<j \le m}$ in $\{0,1\}^{\binom{m}{2}}$ with $X_{(i,j)} = 1$ if there is an edge between $i$ and $j$. Consider the Erdos-Renyi model where $X$ has i.i.d. Bernoulli enties with mean $a \in [0,1]$. We are interested in the concentration of the number of triangles:
$$
f(X) = \sum_{1\le i<j<k \le m} X_{(i,j)} X_{(j,k)} X_{(i,k)}
$$
The expected number of triangles is
$$
\EE(f(X)) = \binom{m}{3} a^3
$$
Let us show that $f$ has bounded differences, consider $X$ and $Y$ such that $X_{(i,j)} = 1$, $Y_{(i,j)} = 0$, and $X_{e} = Y_{e}$ for all $e \ne (i,j)$ and define $\Delta_{(i,j)}$ the discrete derivative of $f$ at $X$ with respect to $X_{(i,j)}$:
$$
	|f(X) - f(Y)| = \Delta_{(i,j)}(X) = \sum_{j < k  \le m} X_{(j,k)} X_{(i,k)}
$$ 
Since $|\Delta_{(i,j)}(X)| \le m-2$ for all $X$, $f$ has $c$-bounded differences with $c=(m-2,...,m-2)$ and McDiarmid's inequality yields:
$$
\PP[f(X) - \EE(f(X)) \geq  \epsilon ] \leq \exp\left( - \frac{2 \epsilon^2}{(m-2)^2 \binom{m}{2}} \right).
$$
This inequality can be substantially improved, because although $\Delta_{(i,j)}(X)$ can be equal to $m-2$ in the worst case, with high probability it will be close to its expectation which can be much smaller than $m$. 

Indeed $\Delta_{(i,j)}(X) \sim$ Binomial$(m-j, a^2)$, so that we can apply Chernoff's inequality in its multiplicative version (see for instance \cite{mitzenmacher}) and a union bound:
\begin{align*}
	\PP( \max_{1 \le i<j \le m} \Delta_{(i,j)}(X) \ge 2 (m-2) a^2)
	&\le \binom{m}{2} (2 e)^{-2(m-2) a^2}
\end{align*}
Therefore, with high probability, $f$ has $c'$-bounded differences with  $c' = (2 (m-2) a^2,...,2 (m-2) a^2)$ and the generalized McDiarmid's inequality yields
$$
\PP[f(X) - \EE( f(X) | X \in \YY) \geq p \bar{c} + \epsilon ] 
$$
$$
\leq p +  \exp\left( - \frac{\epsilon^2}{2 (m-2)^2 a^4 \binom{m}{2}} \right).
$$
with 
$$
	p = \PP(X \not\in \YY) \le \binom{m}{2} (2 e)^{-2(m-2) a^2}
$$
and
\begin{align*}
	\EE( f(X) | X \in \YY) &\le  \EE(f(X)) + p \max_{X \in \XX} f(X) \\
	&= \binom{m}{3} (a^3 +  p)  
\end{align*}
and 
$$
p \bar{c}  \le 2 (m-2) (2 e)^{-2(m-2) a^2} a^2 \left(\binom{m}{2}\right)^2  
$$
We can now compare the two inequalities. The first inequality predicts that with probability greater than $1-\delta$ we have
$$
	f(X) \le  \binom{m}{3} a^3 + \sqrt{ \log(1/\delta) {(m-2)^2 \over 2} \binom{m}{2}} 
$$
while the second inequality predicts that with probability greater than $1-\delta-p$ we have:
$$
f(X) \le  \binom{m}{3} a^3 + p( \binom{m}{3} + \bar{c}) + \sqrt{ \log(1/\delta) 2 (m-2)^2 a^4 \binom{m}{2}}
$$
where $p$ vanishes exponentially with $m$ for any fixed $a$, hence the additional term $p( \binom{m}{3} + \bar{c})$ also vanishes exponentially in $m$. This is a much better bound, especially if $a$ is small. It is also noted that this bound is exponentially better than bounds derived using Chebychev's inequality and similar inequalities based on the first and second moments.

This example gives a blueprint on how to use the generalized inequality: first compute the discrete derivative of $f$ with respect to the entries of $X$, then control the fluctuation of each discrete derivative using a well-chosen concentration inequality, and finally apply a union bound and then the generalized McDiarmid inequality to get the result. It is noted that the above can easily be extended to stochastic block models, and other random graphs where the entries of $X$ are independent but not identically distributed. 

\subsection{Maximum Likelihood Estimation}

Our second example of application is the concentration of the error of the maximum likelihood estimator. Consider $X = (X_1,...,X_n)$ an i.i.d sample drawn from some distribution $p(x|\theta)$ parameterized by $\theta \in \Theta \subset \RR^d$. Define the normalized likelihood of the sample
$$
g(X,\theta) = {1 \over n} \sum_{i=1}^n \log p(X_i|\theta)
 = {1 \over n} \sum_{i=1}^n \ell (X_i,\theta)
$$
Assume that $g$ is $\mu$-strongly concave so that:
$$
	 -\mu I  \succ \nabla_{\theta}^2 g(X,\theta) \text{ for all }  (X,\theta) \in \XX \times \Theta
$$
with $\nabla_{\theta}^2$ the Heissian and $\succ$ the order of symmetric matrices.

We define the maximum likelihood estimator:
$$
f(X) = \arg\max_{\theta \in \Theta} g(X,\theta)
$$
and its error:
$$
h(X) = \| f(X)-\theta \|^2
$$
Let us show that $h$ has bounded differences. Define
$$
	\Delta = \max_{\theta \in \Theta} \max_{(x,y) \in \XX^2} |\ell(x,\theta) - \ell(y,\theta)|
$$
and consider $X$ and $Y$ such that $X_j=Y_j$ for all $j \ne i$.

We have:
\begin{align*}
g(X,f(X)) &- \mu||f(X)-f(Y)||^2 \ge g(X,f(Y)) \\
                &\ge g(Y,f(Y)) - {\Delta \over n} \\
                &\ge g(Y,f(X)) - {\Delta \over n} \\
                &\ge g(X,f(X)) - 2 {\Delta \over n} \\
\end{align*}             
where we used the $\mu$ strong convexity of $g$, the fact that for any $\theta$, $|g(X,\theta) - g(Y,\theta)| \le \Delta/n$, and the fact that $f$ maximizes $g$. We have proven that
$$
	||f(X)-f(Y)||^2 \le {2 \Delta \over n \mu}
$$
and the triangle inequality shows that
$$
	|h(X)-h(Y)| \le ||f(X)-f(Y)|| \le \sqrt{2 \Delta \over n \mu}
$$
So $h$ has $c$ bounded differences with $c = (\sqrt{2 \Delta \over n\mu},...,\sqrt{2 \Delta \over n\mu})$, and McDiarmid's inequality yields:
$$
\PP[h(X) - \EE(h(X)) \geq  \epsilon ] \leq \exp\left( - \frac{\mu \epsilon^2}{\Delta} \right).
$$
While this inequality shows that the error of the maximum likelihood does concentrate around its expected value, it can be substantially improved using the fact that with high probability, $f(X)$ will lie in a neighbourhood of $\theta$.  Consider $\Theta'$ a neigbourhood of $\theta$, and define
$$
	\Delta' = \max_{\theta \in \Theta'} \max_{(x,y) \in \XX^2} |\ell(x,\theta) - \ell(y,\theta)|
$$
Then, from the same logic as above, $h$ has $c'=(\sqrt{2 \Delta' \over n\mu},...,\sqrt{2 \Delta' \over n\mu})$ bounded differences with high probability, and the generalized McDiarmid's inequality yields:
$$
\PP[h(X) - \EE[h(X)| f(X) \in \Theta'] \geq p \bar{c} + \epsilon ] \leq p + \exp\left( - \frac{\mu \epsilon^2}{\Delta'} \right).
$$
with
$$
p = \PP[ f(X) \not\in \Theta']
$$
and
\begin{align*}
	\EE[h(X)|f(X) \in \Theta'] &\le \EE[h(X)] + p \max_{X \in \XX} |h(X)|  \\
	&\le \EE[h(X)] + p {\bf diam}(\Theta).
\end{align*}
and
$$
	p \bar{c} = p \sqrt{2 n \Delta'/\mu}
$$
We can now compare the two inequalities. Typically, $p$ decays exponentially with $n$, so that both $p {\bf diam}(\Theta)$ and $p \bar{c}$ will be negligible. Both inequalities are therefore similar, except for the fact that $\Delta$ has been replaced by $\Delta'$, which is equivalent to replacing $\Theta$ by $\Theta'$. While this may seem inoccuous, in many models the first inequality is uninformative, while the second one is useful. Consider for instance the simple case where $X_i$ has Bernoulli($\theta$) distribution $\theta \in \Theta = [0,1]$. Then 
$$
\ell(x,\theta) = x \log \theta + (1-x) \log (1-\theta)
$$
so that $\ell(x,\theta)$ is $\mu$-strongly concave for $\mu=1$, and $g$ is also $\mu$-strongly concave.

However:
$$
	\Delta = \max_{\theta \in \Theta} \max_{(x,y) \in \XX^2} |\ell(x,\theta) - \ell(y,\theta)| = +\infty
$$
so that McDiarmid's inequality is uninformative, while one the other hand, as long as $\Theta'$ is comprised in the interior of $\Theta$:
\begin{align*}
	\Delta' &= \max_{\theta \in \Theta'} \max_{(x,y) \in \XX^2} |\ell(x,\theta) - \ell(y,\theta)| \\
	&\le \log\left( {\max_{\theta \in \Theta'} \max(\theta,1-\theta) \over \min_{\theta \in \Theta'} \min(\theta,1-\theta)}\right)  < \infty.
\end{align*}
so that the generalized McDiarmid's inequality yields a useful bound.

\subsection{Empirical Risk Minimization}

The previous example can further be extended to study the concentration of empirical risk minimizers. In that case one can replace $-\ell(x,\theta)$ by an arbitrary loss function, providing that this loss function is $\mu$ strongly convex, and is locally bounded. For instance, this is true for regularized risk minimization, where $-\ell(x,\theta) = g(x,\theta) + \mu \| \theta \|^2$ where $\theta \mapsto g(x,\theta)$ is any convex function. See~\cite{shalevshwartz} for an overview of regularized risk minimization.

\section{Extension to general metric spaces}
\subsection{Statement}
Interestingly, our results can be further extended to a more general scenario: concentration of Lipschitz continuous functions on a high probablity subset in a general metric space. For instance, this allows to apply the approach to Gaussian concentration.

Consider $X$ a random variable in a metric space $(\XX,d)$, for any $\YY \subset \XX$ define
$$
	\LL(\YY) = \{ f:\XX \to \RR: |f(x)-f(y)| \le d(x,y) \;,\;\forall (x,y) \in \YY^2\}
$$
the set of $1$-Lipschitz functions over subset $\YY$ and 
$$
	\Phi(\epsilon) = \sup_{f \in \LL(\XX)} \PP[  f(X) - \EE[f(X)] \ge \epsilon]
$$
a function which controls the concentration of Lipschitz functions over $\XX$ around their expectation.  As a preliminary we recall the definition of the Wasserstein distance, which is instrumental to our results.
\begin{definition}\label{def:wasserstein}
The Wasserstein distance between distributions $\mu$ and $\nu$ is defined as:
$$
	W_1(\mu,\nu)  = \inf\left\{ \EE[ d(X,Y) ] \text{ with } X \sim \mu \text{ and } Y \sim \nu \right\}
$$
where the infimum is taken over all the possible couplings $(X,Y)$ with marginal distributions $\mu$ and $\nu$.
\end{definition}

\begin{proposition}\label{prop:mcdiarmid_metric}
Consider a subset $\YY \subset \XX$, $\YY^c = \XX \setminus \YY$ its complement and a function $f \in \LL(\YY)$. 

Then for all $\epsilon \ge 0$ we have
$$
	\PP[ f(X) - \EE[f(X) | X \in \YY] \ge \epsilon + p W_1(P_{X | \YY},P_{X | \YY^c})] \le \Phi(\epsilon)
$$
with $p = \PP[X \not\in \YY]$, $P_{X | \YY}$ and $P_{X | \YY^c}$ the distributions of $X$ conditional to $X \in \YY$ and $X \in \YY^c$ respectively.
\end{proposition}
Proposition~\ref{prop:mcdiarmid_metric} states that if Lipschitz functions concentrate around their expectation, then Lipschitz functions on a high probability subset also concentrate around their conditional expectation, up to an error term proportional to the Wasserstein distance between $P_{X | \YY}$ and $P_{X | \YY^c}$. This is interesting, as there are strong links between the Wasserstein distance and Lipschitz functions due to the Kantorovich duality. The error term can be controlled under mild assumptions if $\YY$ is a high probability set, as shown in two corollaries, even when the metric space is unbounded, providing that the distribution of $X$ has bounded second moment.
\begin{corollary}
For any $(\XX,d)$ we have the upper bound:
$$
	p W_1(P_{X | \YY},P_{X | \YY^c}) \le \sqrt{p/(1-p)} \sqrt{\EE[d(X,X')^2]}
$$
where $X$ and $X'$ are i.i.d. with distribution $P_X$.
\end{corollary}

\begin{corollary}
For $(\XX,d)$ with a finite diameter we have the upper bound:
$$
	p W_1(P_{X | \YY},P_{X | \YY^c}) \le p \sup_{(x,y) \in \XX^2} d(x,y)  
$$
\end{corollary}

We now provide two illustrative examples, showing that proposition~\ref{prop:mcdiarmid_metric} includes proposition~\ref{prop:main} as a particular case, and also that it is readily applicable to Gaussian concentration. 
\begin{example} 
	Consider $c \in (\RR^{+})^n$, the metric space $(\XX,d_c)$ with $d_c$ the weighted Hamming distance $d_c(x,y) = \sum_{i=1}^n c_i \indic(x_i \ne y_i)$. We have $f \in \LL(\XX)$ if $f$ has $c$-bounded differences, and from McDiarmid's inequality
$$\Phi(\epsilon) = \sup_{f \in \LL(\XX)} \PP[  f(X) - \EE[f(X)] \ge \epsilon] \le \exp\left( - \frac{2 \epsilon^2}{\sum_{i=1}^n c_i^2} \right)$$ We also note that the diameter of $(\XX,d_c)$ is bounded by $\bar{c} = \sum_{i=1}^n c_i$. So in that case proposition \ref{prop:mcdiarmid_metric} reduces to proposition \ref{prop:main}.   
\end{example}

\begin{example} 
	Consider metric space $(\XX,d)$ with $\XX=\RR^n$, $d$ the Euclidan distance and $X \sim N(0,I_n)$ a standard Gaussian vector. As shown in \cite{ledoux} for all $\epsilon \ge 0$:
$$\Phi(\epsilon) = \sup_{f \in \LL(\XX)} \PP[  f(X) - \EE[f(X)] \ge \epsilon] \le \exp\left(- {\epsilon^2 \over 2} \right)$$
Hence for any $1$-Lipschitz function $f$ over $\YY \subset \XX$, for all $\epsilon \ge 0$ we have
$$
	\PP[  f(X) - \EE[f(X) | X \in \YY] \ge \epsilon + \sqrt{2 n p / (1 - p)}  ] \le \exp\left(- {\epsilon^2 \over 2} \right)
$$
since $\sqrt{\EE[d(X,X')^2]} = \sqrt{2 n}$.
\end{example}

\subsection{Proof}
The proof follows similar arguments as the previous one, once again, since $f \in \LL(\YY)$, one can use the extension technique of \cite{mcshane},
$$
g(x) = \inf_{y \in \YY} \{  f(y) + d(x,y) \}.
$$
where $g$ is the Lipschitz extension of $f$: $g \in \LL(\XX)$ and $g(x) = f(x)$ for all $x \in \YY$. Decomposing:
\als{
\PP[ f(X) - \EE[g(X)] \geq \epsilon] &\leq  \PP[ f(X) - \EE[g(X)] \geq \epsilon , X \in \YY] \\
&+ \PP[X \not\in \YY]
}
and since $g$ is the Lipschitz extension of $f$
\als{
\PP[ f(X) - \EE[g(X)] \geq \epsilon , X \in \YY] &\le \PP[g(X) - \EE[g(X)] \geq \epsilon] \\
&\le \Phi(\epsilon)
}
To bound the expectation of $g$, consider any $(Z_1,Z_2)$ with marginals $P_{X | \YY}$ and $P_{X | \YY^c}$ and let 
$$
	(X,Y) = \begin{cases} 
			(Z_1,Z_1) &\text{ with probability } 1-p \\
			(Z_2,Z_1) &\text{ with probability } p 
		\end{cases}
$$
It is noted that $X \sim P_X$. We then have 
$$
	g(X) = \inf_{y \in \YY} \{  f(y) + d(X,y) \} \le f(Y) + d(X,Y)  
$$
Taking expectations:
$$
	\EE[ g(X) ] \le \EE[f(Z_1)] + p \EE[d(Z_1,Z_2)]  
$$
The above holds for any $(Z_1,Z_2)$ with marginals $P_{X | \YY}$ and $P_{X | \YY^c}$ so by definition of the Wasserstein distance:
$$
	\EE[ g(X) ] \le \EE[f(X)|X \in \YY] + p W_1(P_{X | \YY},P_{X | \YY^c})
$$
yielding the proposition.

For any $(\XX,d)$, consider $X,X'$ i.i.d. with distribution $P_X$ and:
\begin{align*}
W_1(P_{X | \YY},P_{X | \YY^c}) &\le \EE[ d(X,X') | (X,X') \in \YY \times \YY^c] \\ 
&= {\EE[ \indic((X,X') \in \YY \times \YY^c) d(X,X')] \over \PP[(X,X') \in \YY \times \YY^c]}  
\end{align*}
The Cauchy-Schwartz inequality gives
\als{
	\EE[& \indic((X,X') \in \YY \times \YY^c) d(X,X')] \\\ &\le \sqrt{ \PP[(X,X') \in \YY \times \YY^c]} \sqrt{\EE[ d(X,X')^2]}
}
From independence 
$$\PP[(X,X') \in \YY \times \YY^c] = p(1-p)$$
and replacing 
$$
	p W_1(P_{X | \YY},P_{X | \YY^c}) \le \sqrt{p/(1-p)} \sqrt{\EE[ d(X,X')^2]}
$$
yielding the first corollary.

If $(\XX,d)$ has finite diameter for any $(X,Y)$ with marginals $P_{X | \YY}$ and $P_{X | \YY^c}$:
$$
p W_1(P_{X | \YY},P_{X | \YY^c}) \le p \EE[d(X,Y)] \le p \sup_{(x,y) \in \XX^2} d(x,y) 
$$
yielding the second corollary.

\section{Conclusion}

We have generalized McDiarmid's inequality for functions with bounded differences on a high probability set and demonstrated the usefulness of this generalization on several examples taken from random graphs, statistics and learning theory. In those examples, the generalized inequality yields considerably tighter bounds, without much additional complexity. As a direction for future work, it would be interesting to determine in which cases this generalized inequality is tight.

\newpage
\bibliographystyle{plain}
\bibliography{mcdiarmid}
\end{document}